\ifdefined\XeTeXversion\else
  \pdfoutput=1
\fi
\documentclass[11pt]{article}

\PassOptionsToPackage{table}{xcolor}
\PassOptionsToPackage{hyperfootnotes=false}{hyperref}
\usepackage[preprint]{acl}

\usepackage{times}
\usepackage{latexsym}

\usepackage[T1]{fontenc}

\usepackage[utf8]{inputenc}

\usepackage{microtype}

\usepackage{inconsolata}

\usepackage{graphicx}

\usepackage{amssymb}     
\usepackage{pgfplots}
\usepackage{pgf-pie}     
\usepackage{booktabs}    
\usepackage{colortbl}    
\pgfplotsset{compat=1.18}
\usepgfplotslibrary{fillbetween}  
\usetikzlibrary{positioning, arrows.meta, calc, fit, backgrounds, shapes.geometric}

\AtBeginDocument{\hypersetup{breaklinks=false}}

\newcommand{\evalN}{216}        
\newcommand{\nModels}{9}        
\newcommand{\agentRounds}{3}    

\newcommand{\corpusSubs}{220{,}943}         
\newcommand{\corpusComments}{18{,}094{,}365}
\newcommand{\corpusDocs}{18.3M}             
\newcommand{\corpusMonths}{25}              
\newcommand{\corpusSubreddits}{20}          
\newcommand{\corpusSpan}{January 2024 through January 2026}

\newcommand{\poolEvents}{1{,}916}     
\newcommand{\poolMarkets}{7{,}913}    
\newcommand{\poolResolved}{3{,}427}   
\newcommand{\probeK}{5}               

\newcommand{\sysname}{\textsc{Hindcast}}

\title{\sysname{}: Replaying Prediction Markets to Evaluate LLM Forecasters}

\author{
    Xiao Ye\thanks{\;\, Equal contribution.} \quad
    Jacob Dineen\footnotemark[1] \quad
    Evan Zhu \quad Shijie Lu \quad Kevin Song \quad Ben Zhou \\
    School of Computing and Augmented Intelligence, Arizona State University \\
    \texttt{\{xiaoye2, jdineen\}@asu.edu}
  }

\begin{document}
\maketitle
\begin{abstract}
Forecasters are evaluated by backtesting, which replays resolved questions and grades the probability the system would have assigned before the outcome was known. For LLMs, two channels leak the answer into this test. A model that retrieves can surface reports written after the event, turning forecasting into a lookup, and each new model is trained on data closer to the event, so a question that lay in the future for last year's models sits inside this year's training data. Either way, the test grades recall while claiming to grade foresight. We introduce \sysname{}, which closes both leaks by grading a model as if it stood at a chosen past date $t_0$, before the outcome existed in either channel. \sysname{} replays resolved Polymarket prediction markets against a frozen snapshot of public Reddit, lets the model read only posts written before $t_0$, and scores each forecast against both what happened and the market's own price at $t_0$, itself a human forecast made from the same past information. Because the cutoff is set per market and the snapshot never changes, the evaluation re-runs on new markets as models improve, without going stale. Once the leak is closed, retrieval still helps most models, but only where Reddit discussed the event beforehand. Where the archive carried only speculation, retrieval hurts.
\end{abstract}

\section{Introduction}
\label{sec:intro}

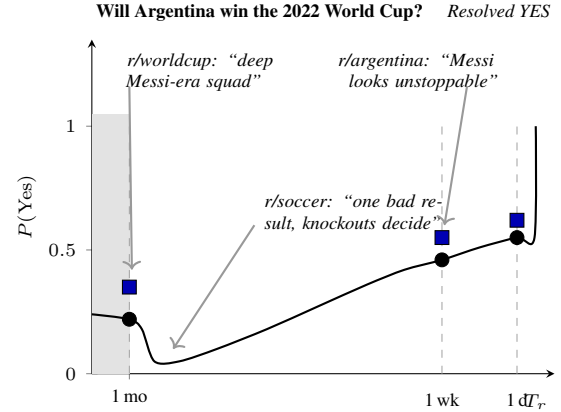
\begin{figure}[!t]
  \centering
  \begin{tikzpicture}
    \begin{axis}[
      width=\columnwidth,
      height=6.0cm,
      xmin=0, xmax=37,
      ymin=0, ymax=1.35,
      xtick={3},
      xticklabels={1\,mo},
      xticklabel style={font=\scriptsize},
      ytick={0, 0.5, 1},
      yticklabel style={font=\scriptsize},
      ylabel={$P(\mathrm{Yes})$},
      ylabel style={font=\scriptsize, yshift=-1mm},
      tick align=outside,
      axis lines=left,
      title={\scriptsize\textbf{Will Argentina win the 2022 World Cup?}
             \quad{\scriptsize\itshape Resolved YES}},
      title style={yshift=-1mm},
      clip=false,
    ]
      \fill[gray!22] (axis cs:0,0) rectangle (axis cs:3, 1.05);

      \addplot[thick, black, smooth, mark=none] coordinates {
        (0,0.24) (3,0.22) (4,0.18) (5,0.05) (7,0.05) (10,0.10)
        (14,0.18) (18,0.27) (22,0.36) (25,0.42) (28,0.46)
        (31,0.51) (34,0.55) (35.4,0.55) (35.5,1.0)
      };

      \draw[dashed, gray!70] (axis cs:3,0)  -- (axis cs:3,1);
      \draw[dashed, gray!70] (axis cs:28,0) -- (axis cs:28,1);
      \draw[dashed, gray!70] (axis cs:34,0) -- (axis cs:34,1);

      \addplot[only marks, mark=*, mark size=2.6pt, black]
        coordinates {(3,0.22) (28,0.46) (34,0.55)};

      \addplot[only marks, mark=square*, mark size=2.6pt,
               fill=blue!70!black, draw=black, line width=0.4pt]
        coordinates {(3,0.35) (28,0.55) (34,0.62)};

      \node[font=\scriptsize, anchor=north] at (axis cs:28, -0.04) {1\,wk};
      \node[font=\scriptsize, anchor=north] at (axis cs:34, -0.04) {1\,d};
      \node[font=\scriptsize, anchor=north] at (axis cs:35.5, -0.04) {$T_r$};

      \node[font=\scriptsize, align=left, anchor=west, text width=2.3cm]
        at (axis cs:2, 1.22)
        {\itshape r/worldcup: ``deep Messi-era squad''};
      \draw[->, thick, gray!80]
        (axis cs:3, 1.16) -- (axis cs:3.2, 0.42);

      \node[font=\scriptsize, align=left, anchor=west, text width=2.5cm]
        at (axis cs:13, 0.65)
        {\itshape r/soccer: ``one bad result, knockouts decide''};
      \draw[->, thick, gray!80]
        (axis cs:13, 0.60) -- (axis cs:6.5, 0.10);

      \node[font=\scriptsize, align=right, anchor=east, text width=2.4cm]
        at (axis cs:33, 1.22)
        {\itshape r/argentina: ``Messi looks unstoppable''};
      \draw[->, thick, gray!80]
        (axis cs:30, 1.16) -- (axis cs:28.2, 0.60);
    \end{axis}
  \end{tikzpicture}
  \caption{\textbf{\sysname{} on a resolved market (illustrative).}
    At a simulated query time $t_0$ (a one-month lookback here, pinned
    per market in our experiments, \S\ref{sec:eval-metrics}), the
    agent sees only Reddit content created before $t_0$ and emits a
    forecast (\textcolor{blue!70!black}{$\blacksquare$}), scored
    against the market's implied probability at the same $t_0$
    ($\bullet$). Argentina's group-stage loss to Saudi Arabia drops
    the market sharply, while the agent's forecast tracks Reddit's
    steadier tournament-long narrative.}
  \label{fig:teaser}
\end{figure}

Calibrated probabilistic forecasting is becoming a standard test of whether a language model can reason under uncertainty, and increasingly a way to decide which agents to trust with open-ended tasks~\citep{ye2025evaluating}. Writing forecasting questions by hand is slow, so the common shortcut is to grade models on \emph{resolved} questions whose answers are now known and to score the probability the model assigns to the true outcome~\citep{halawi2024approaching,ye2024mirai}. The supervision is automatic and cheap, but it pays a model for memory just as readily as for foresight. A model asked today for the 2022 World Cup champion (Figure~\ref{fig:teaser}) names Argentina with near-certainty, and that certainty comes from the post-tournament write-ups in its training data. At inference, a model that retrieves can pull in content created after the event, and across releases each new model is trained on data closer to the event, so a question safely out-of-cutoff for this year's models leaks into next year's~\citep{yan2026datedgpt,levine2026talkie}. Such an evaluation measures how well a model recalls the resolved outcome, while its foresight goes untested.

What we actually want to measure is the counterfactual probability the model \emph{would} have assigned at a past time $t_0$, before the outcome was known. This is also the only forecast that was ever useful, since it is the one a trader or analyst could have acted on at $t_0$~\citep{tetlock2015superforecasting}. But it is hard to recover after the fact, because the live web has moved on. Pages are rewritten and re-indexed, so retrieval can no longer return what was actually visible at $t_0$. Live benchmarks~\citep{karger2025forecastbench,zeng2025futurex} avoid this by posing questions that are still open and waiting for them to resolve, but waiting forfeits the two things resolved questions provide together. A resolved question gives a known outcome to grade against, and a prediction market adds a human forecast at $t_0$ to measure against. That tension sets up the question this paper answers.

\begin{quote}
\emph{On already-resolved questions, can we evaluate the forecast a model would have made at a past $t_0$, holding fixed the information it could retrieve then?}
\end{quote}

We answer with \sysname{}, which makes this counterfactual measurable through \emph{hindcasting}, by replaying each resolved market against a frozen archive of the past and pinning every retrieval call to that market's $t_0$. Two design choices follow directly. First, the archive must not change after the fact, so we build it from public Reddit, whose monthly snapshots are immutable (unlike web pages or Wikipedia, a captured month stays captured) and which discusses unfolding events early and broadly, giving a forecaster real pre-event signal to find. Second, the questions need a yardstick, so we draw on resolved Polymarket markets, each of which supplies the graded outcome and a matched-time price at $t_0$ that is itself a human forecast distilled by traders from the same public information. Because the past is fixed per market, $t_0$ becomes a free variable and the evaluation re-runs on markets that resolve after a new model ships, so it never decays into recall. The concurrent FutureSim~\citep{goel2026futuresimreplayingworldevents} shares the replay idea but takes a different shape. It streams a news corpus forward over a single shared window to measure adaptation, while \sysname{} pins an immutable Reddit archive to a per-market $t_0$ and scores against the matched-time market price.

\begin{figure*}[t]
  \centering
  \includegraphics[width=\textwidth]{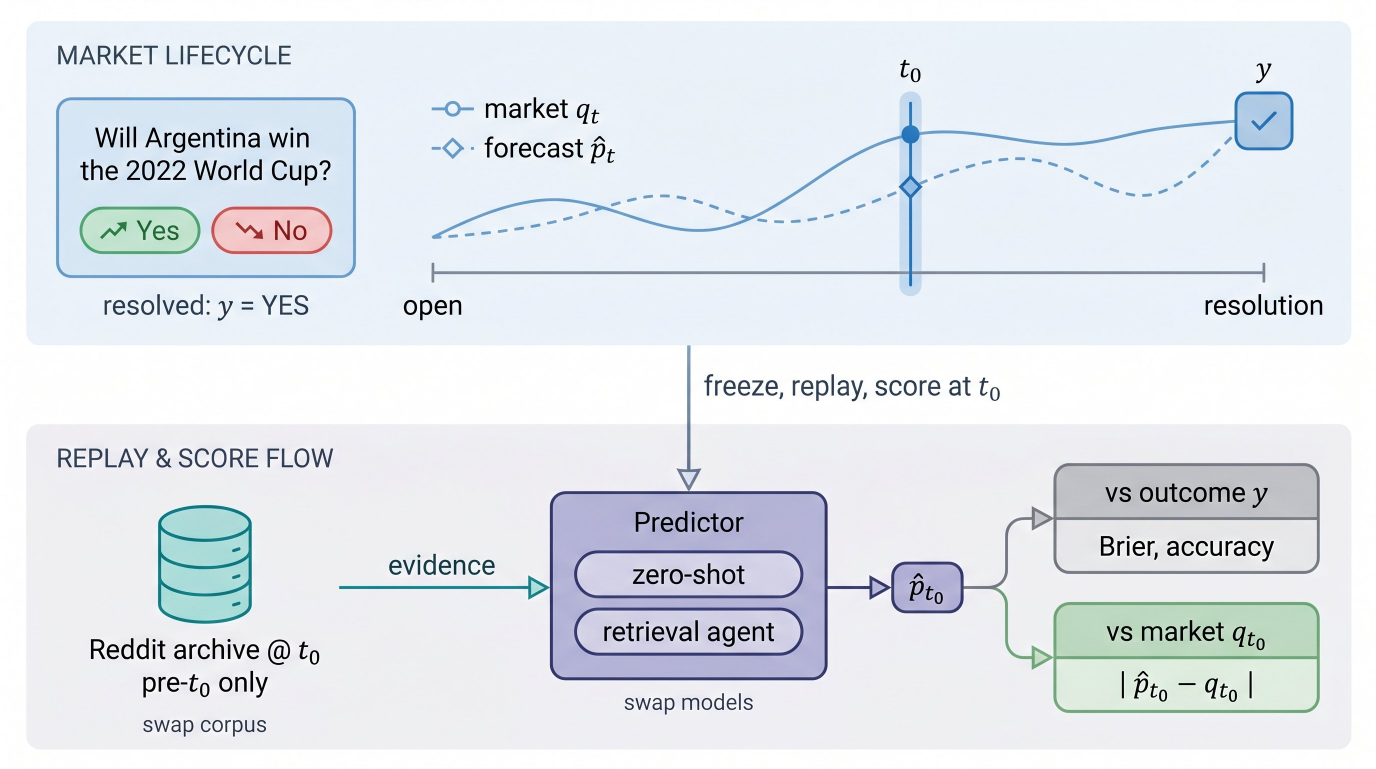}
  \caption{\textbf{Overview of \sysname{}.} Any resolved Polymarket
    market can be frozen at any past time $t_0$ along its open-to-resolution
    lifecycle (top). At the chosen $t_0$, the Reddit archive is restricted
    to documents created before $t_0$, and a predictor (zero-shot baseline
    or retrieval-grounded agent) reads that frozen evidence and emits
    a probability $\hat{p}_{t_0}$. The forecast is scored against both
    the resolved outcome $y$ (Brier, accuracy) and the contemporaneous
    market-implied probability $q_{t_0}$ ($|\hat{p}_{t_0}-q_{t_0}|$).
    Both the corpus and the predictor can be swapped without changing
    the protocol, and sweeping $t_0$ along the lifecycle yields a
    calibration trajectory for any benchmarked model.}
  \label{fig:system}
\end{figure*}

Both predictors sit behind one interface that takes a market, its resolution criteria, and a cutoff $t_0$, and returns a probability with a short rationale (\S\ref{sec:pipelines}). We instantiate it twice. The \emph{zero-shot baseline} forecasts from the model alone and isolates parametric memory, while the \emph{Retrieval agent} adds only access to the archive at $t_0$, so the gap between them measures the contribution of evidence. Figure~\ref{fig:system} sketches the pipeline.

With a leakage-free evaluation in hand, we revisit whether retrieval-grounded reasoning improves forecasting, a claim prior work makes under possible contamination. If those reported gains came partly from reading the future, they should shrink once the future is sealed off. The gain does not vanish, since retrieval still lowers Brier on eight of the nine open-weight models we test. But it concentrates on markets the corpus covered in advance and reverses where the agent over-reads speculative chatter (\S\ref{sec:analysis}). Retrieval helps, then, but only when the pre-$t_0$ corpus already discussed what decided the event. Where it carried only speculation, retrieval backfires.

\paragraph{Contributions.}
(i)~\emph{Hindcasting}, an evaluation protocol that measures a model's forecast at a past $t_0$ on resolved questions, holding its retrievable information fixed, and that re-runs as models advance (\S\ref{sec:backend},~\S\ref{sec:eval}). (ii)~An immutable, temporally pinned public-Reddit archive serving leakage-free retrieval at any past $t_0$ (\S\ref{sec:backend}). (iii)~Two reference predictors behind one probability-emitting interface, a zero-shot baseline and a Retrieval agent (\S\ref{sec:pipelines}). (iv)~An empirical study locating where retrieval-grounded forecasting helps and where it backfires (\S\ref{sec:experiments}). What limits such a forecaster turns out to be its evidence environment.

\section{Related Work}
\label{sec:related}

\paragraph{LLM forecasters and benchmarks.}
Early benchmarks frame future-event prediction as temporally restricted QA over news or tournament questions \citep{jin2021forecastqa,zou2022forecasting}. Retrieval-augmented and agentic systems push this further, into open-ended forecasting against eventual outcomes \citep{halawi2024approaching,ye2024mirai,das2026nexusagenticframework}. Concurrent with our work, FutureSim~\citep{goel2026futuresimreplayingworldevents} streams news forward chronologically over a recent window to measure adaptation. \sysname{} instead holds the corpus fixed at a chosen $t_0$ per market and measures improvement over the matched-time market price. Live benchmarks counter static-test contamination through continual collection \citep{karger2025forecastbench,zeng2025futurex,yang2025llm}. A parallel line moves closer to prediction-market deployment, casting the task as trading over live market states \citep{yu2025livetradebench,cheng2026polybench}. \sysname{} is the inverse move, replacing online evaluation with replay against a fixed past information environment, so any predictor can be re-run against any past $t_0$ without waiting for new questions to resolve.

\paragraph{Temporal controls and contamination.}
Existing responses to contamination refresh evaluation items \citep{vu2024freshllms,white2024livebench,jain2024livecodebench,benchmark_contamination_survey}, retrain models from scratch on temporally-restricted corpora that are either annually partitioned \citep{yan2026datedgpt} or pinned to a single pre-cutoff vintage \citep{levine2026talkie}, or fix the retrieval context to a static news snapshot \citep{chandak2025scaling,nagel2016ccnews,cheng2024dateddata}. The retraining strategies attack the parametric-memory channel by removing post-cutoff text from the weights, whereas \sysname{} attacks the retrieval channel by pinning the archive at $t_0$, so the two approaches are complementary. Archived Reddit at the simulated query time gives the agent a denser, more topically varied source of information than a single news snapshot can provide.

\paragraph{Prediction markets and calibration.}
Prediction markets supply resolvable questions, crowd probabilities, and tradable payoffs. Prior work has used them as evaluation targets \citep{zou2022forecasting,halawi2024approaching,karger2025forecastbench}, as delayed supervision under proper scoring rules \citep{turtel2026future}, and as trading environments measured by market returns \citep{yang2025llm,yu2025livetradebench,cheng2026polybench}. We use Polymarket as a dual benchmark. Resolved binary markets supply ground truth, and historical market-implied probabilities supply a matched-time baseline. Together they let us separately measure outcome recovery, improvement over the market's information state at $t_0$, and calibration as resolution approaches. We score calibration directly under proper rules such as the Brier score \citep{glenn1950verification,gneiting2007strictly}. Methods for sharpening that calibration and the reasoning behind it range from LLM ensembling \citep{schoenegger2402wisdom} and outcome-based training \citep{turtel2026future,chandak2025scaling} to Bayesian and inference-time factor decomposition \citep{feng2024bird,srinivasan2026recaptransparentinferencetimeemotion} and consistency training \citep{ye2025cclearncohortbasedconsistencylearning}. \sysname{} is a controlled testbed under which such retrieval and calibration strategies can be replayed at matched timepoints and compared against both binary outcomes and contemporaneous market probabilities.

\section{Retrieval Backend}
\label{sec:backend}

The information side of \sysname{} is served by a temporally pinned archive of public Reddit. The backend keeps one immutable copy of each document, indexes it for retrieval, and answers every call under a single cutoff predicate. It does not re-rank documents, apply a learned post-filter, or judge which results are relevant. Any difference between predictors therefore stems from how the agent queries and reads the archive.

\subsection{Corpus}
\label{sec:backend-corpus}

We build the corpus from the Pushshift archive of public Reddit submissions and comments~\citep{baumgartner2020pushshift}, which provides static monthly dumps captured at posting time. The archive spans \corpusMonths{} consecutive monthly snapshots, \corpusSpan{}, drawn from \corpusSubreddits{} subreddits that together cover the topics our markets touch.

Filtering keeps content that carries an identifiable author and enough engagement to be worth reading, and drops the rest. A record is dropped if it lacks a stable id, if its author or body is deleted or removed, or if it matches an optional bot list. We also drop records below our engagement and length floors, which cut any document with a voting score below $5$, any submission with fewer than $3$ comments, any comment shorter than $5$ non-URL characters or consisting mostly of a link, and any document with fewer than $20$ whitespace-delimited words. Since the filters do not condition on subreddit, topic, or stance, the corpus stays uncurated by content. We keep it broad on purpose, since the eval set spans markets across politics, sports, crypto, macroeconomics, and entertainment, and a topic-narrow corpus would advantage predictors on the topics it covers and penalize them everywhere else.

After filtering, the archive holds \corpusSubs{} submissions and \corpusComments{} comments ($\approx\corpusDocs{}$ documents). Each document is stored once with its text and metadata (subreddit, author, a coarse month key, and an exact creation timestamp), and the snapshot does not change between $t_0$ and evaluation even when the post is later edited or removed on the live site.

\subsection{Retrieval}
\label{sec:backend-search}

The corpus is loaded into PostgreSQL with a \texttt{pgvector} embedding column and is reached two ways, through dense semantic search and through direct by-id lookup. For dense search the query is embedded with the Qwen3-Embedding-0.6B encoder of \citet{zhang2025qwen3embedding} and documents are ranked by cosine distance over per-month HNSW indexes~\citep{malkov2020hnsw}. The reference agent restricts this search to submissions through a \texttt{doc\_type} filter, so every hit is a whole post. This keeps the retrieved set focused, avoids burying the signal under noisy, near-duplicate comment chatter, and returns self-contained units the agent can reason over. Submissions also carry the title and self-text that best summarize a thread's topic.

The dense search excludes comments, but the archive still holds all \corpusComments{} of them and exposes them through by-id lookup tools that resolve a post or comment, walk a submission's comment thread, and list an author's prior posts and comments. A predictor can therefore start from a retrieved submission and pull the surrounding discussion for detail the post text alone does not carry. Every call, whether dense search or a by-id lookup, carries the same filter block. That block holds the cutoff $t_0$, the document type, and optional subreddit, author, month, and start-time restrictions, so the agent can narrow a query without ever weakening the cutoff guarantee. The full tool surface and signatures are deferred to Appendix~\ref{sec:appendix-tools}.

\subsection{Cutoff enforcement}
\label{sec:backend-cutoff}

To make the cutoff guarantee robust to routing errors, we enforce it at two levels. The primary guarantee is a \emph{row-level predicate} under which every document a tool returns is checked against $t_0$ with the strict comparison $\texttt{created\_at} < t_0$, evaluated in the SQL \texttt{WHERE} clause of dense search and of all four by-id lookup tools (post, comment, thread, and author). On top of this, a \emph{partition selector} guards and accelerates dense search by deriving from $t_0$ the monthly partitions no later than $t_0$'s month (and no earlier than any start-time bound), running a separate top-$k$ search against each partition's month-scoped HNSW index, and merging the per-partition results into an exact global top-$k$.

A leak through dense search would thus require both layers to fail on the same document, and a leak through a direct lookup would require the row-level predicate to fail. The boundary partition (the cutoff month itself) is searched, but its post-$t_0$ rows are removed by the row-level predicate. Latency measurements against the live index are in Appendix~\ref{sec:appendix-latency}. 

\section{Evaluation Setup}
\label{sec:eval}

\subsection{Eval-set construction}
\label{sec:eval-construction}

A resolved market tests a retrieval-grounded predictor only if the Reddit corpus discussed its event before it resolved, so resolution alone is not a sufficient inclusion criterion. The eval set must also avoid two skews that would otherwise let aggregate accuracy be driven by a constant guess or a single domain, since most Polymarket markets resolve \textsc{No} and their topics are long-tailed. We therefore construct the eval set in two stages, first a coverage test that keeps only well-supported markets, then outcome and topic balancing.

\paragraph{Coverage test.}
The candidate pool is the \poolMarkets{} binary markets in our Polymarket window (across \poolEvents{} events), of which \poolResolved{} resolve to a Yes/No outcome. For each, an automated probe queries the archive at the market's close date and scores how strongly the pre-event corpus supports the resolved side. An LLM judge (Qwen3-32B) issues $\probeK{}$ queries per side and labels each returned block on a four-level scheme. The label \textsc{strong} (weight $+2$) marks explicit, specific evidence for the resolved side, \textsc{weak} ($+1$) marks vague, indirect, or correlational evidence, \textsc{irrel} ($0$) marks blocks that are on-topic but do not bear on the hypothesis, and \textsc{contra} ($-1$) marks evidence against the resolved side. Per side this yields a score $s_{\mathrm{side}} = 2\,\mathrm{strong} + \mathrm{weak} - \mathrm{contra}$. A final step reads both side tallies together and emits a contrastive coverage score $c \in [-1, 1]$, positive when the corpus supports the resolved side, alongside an overall evidence-quality label of \textsc{strong}, \textsc{weak}, or \textsc{none}. A market the corpus barely mentions scores near zero on both sides and lands on \textsc{none}, so it cannot exercise the retrieval pipeline. We keep only the markets the probe labels \textsc{strong}.

\begin{figure}[t]
\centering
\definecolor{pieSports}{HTML}{4E79A7}
\definecolor{pieTrading}{HTML}{F28E2B}
\definecolor{pieElections}{HTML}{59A14F}
\definecolor{pieAwards}{HTML}{E15759}
\definecolor{pieEnt}{HTML}{B07AA1}
\definecolor{pieTech}{HTML}{76B7B2}
\definecolor{pieCrypto}{HTML}{EDC948}
\definecolor{pieOther}{HTML}{BAB0AC}
{\small
\begin{tikzpicture}
\pie[text=legend,
radius=1.9,
sum=auto,
after number=\%,
rotate=90,
color={pieSports, pieTrading, pieElections, pieAwards,
pieEnt, pieTech, pieCrypto, pieOther}]{
  26.4/Sports (57),
  23.1/Trading (50),
  13.0/Elections (28),
  6.5/Awards (14),
  6.0/Entertainment (13),
  5.1/Tech \& AI (11),
  4.2/Crypto (9),
  15.7/Other (34)
}
\end{tikzpicture}
}
\caption{Topic distribution of the \evalN{}-market eval set. Each slice is labeled with its percent of the set, and the legend gives the raw market count per topic. \emph{Other} aggregates the smaller topics (Social, Geopolitics, Economic, Health, Corporate, and Weather). Topics are unconstrained and follow the markets that carry \textsc{strong} pre-event coverage, while outcomes are balanced by construction.}
\label{fig:eval-composition}
\end{figure}
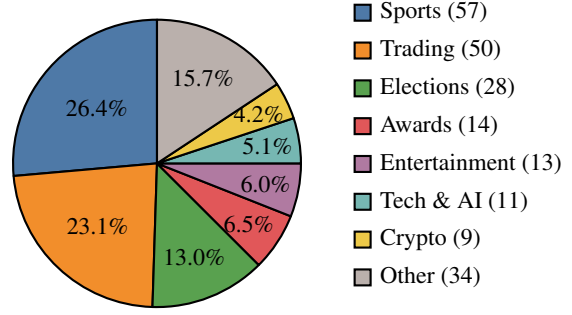

\paragraph{Outcome and topic balancing.}
The strongly-covered markets inherit the skews of the raw pool, since $83\%$ of resolved markets are \textsc{No} ($2{,}851$ of \poolResolved{}) and coverage is uneven across domains because Reddit discusses sports and trading far more than, say, corporate or weather events. We therefore subsample the strongly-covered markets to equalize the Yes/No outcomes and to spread topics as evenly as the available evidence allows, prioritizing outcome balance over topic uniformity. This leaves \evalN{} markets ($106$ \textsc{Yes} and $110$ \textsc{No}), split $36$/$129$/$51$ across short, medium, and long durations, and Figure~\ref{fig:eval-composition} shows the resulting topic distribution. The probe and this selection run as a single automated pipeline that we release, so the eval set can be regenerated as new markets resolve.

\subsection{Metrics and cutoff}
\label{sec:eval-metrics}

Each predictor returns a Yes-side probability $\hat{p} \in [0, 1]$ and a chosen option $\hat{y} = \mathbb{1}[\hat{p} > 0.5]$. We score against the resolved binary outcome $y \in \{0, 1\}$ with accuracy and the Brier score,
\begin{equation}
  \mathrm{Acc} \;=\; \mathbb{1}\!\bigl[\hat{y} = y\bigr],
  \qquad
  \mathrm{Brier} \;=\; (\hat{p} - y)^{2}.
\end{equation}
$\mathrm{Acc}$ rewards a correct decision and $\mathrm{Brier} \in [0,1]$ jointly penalizes miscalibration and miscalls, each averaged over five consistency runs per (model, pipeline) cell. We pin the cutoff $t_0$ to the day before each market closes, capped at the archive's final day (2026-01-31).

\section{Agentic Prediction Pipelines}
\label{sec:pipelines}

\begin{figure}[t]
  \centering
  \includegraphics[width=\linewidth]{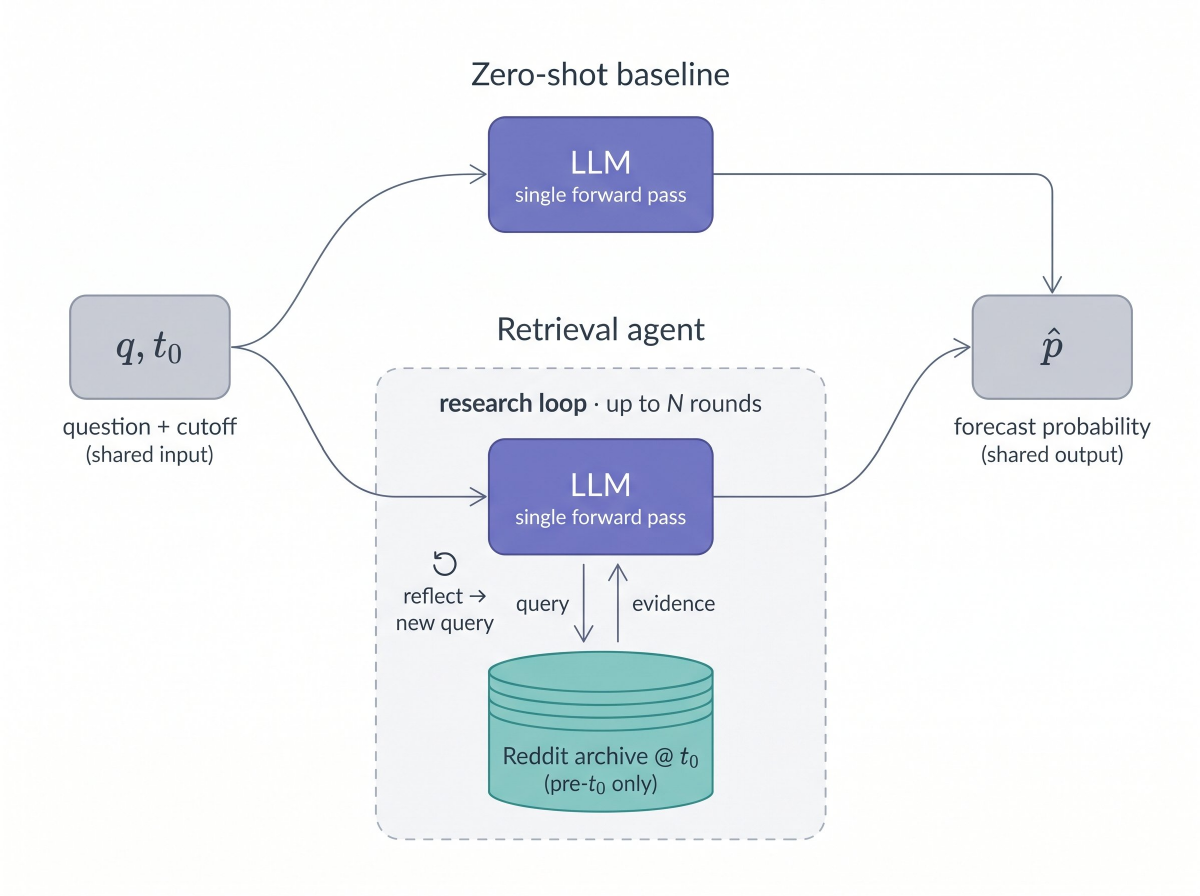}
  \caption{\textbf{The two predictor pipelines, side by side.} The
    zero-shot baseline is a single LLM call. The Retrieval agent runs
    a bounded loop over the archive at $t_0$, alternating queries to
    the archive with a short reflection that proposes the next query,
    then emits a final probability after up to $N$ rounds. Inputs
    ($q$, $t_0$) and the probability-emitting interface are shared,
    and only the body of the call differs.}
  \label{fig:pipelines-compare}
\vspace{-1\baselineskip}
\end{figure}

Both predictor families share an interface, which takes a market, its resolution criteria, an option set, and a cutoff $t_0$. The interface returns a chosen option, per-option probabilities, a self-reported confidence, and a brief rationale. Figure~\ref{fig:pipelines-compare} sketches the two pipelines next to each other.

\subsection{Zero-shot baseline}
\label{sec:pipelines-vanilla}

The zero-shot baseline forecasts from a single LLM call with no retrieval, instructed to use only knowledge that could be known at $t_0$. It sets the parametric-memory floor against which the Retrieval agent is measured.

\subsection{Retrieval agent}
\label{sec:pipelines-agent}

The Retrieval agent runs a bounded research loop over the archive at the $t_0$ cutoff. It opens with seed queries targeting the market's claim and the factors that plausibly drive it, then expands promising hits into their comment threads and author histories. In subsequent rounds, the model reads the accumulated evidence, writes a short reflection that estimates its confidence and names the evidence it still needs, and proposes new queries. The loop terminates on a confidence signal, on diminishing query yield, or on a round budget. A final forecast pass reads the labeled evidence under the same four-level scheme used by the coverage probe (\S\ref{sec:eval-construction}) and emits the same probability format as the zero-shot baseline. The two families therefore differ only in whether the archive is consulted. Cutoff enforcement lives in the backend (\S\ref{sec:backend-cutoff}), so the agent never needs to be trusted with it.

Both pipelines forecast at temperature $0.1$ with top-$p$ $1.0$, and the agent raises the temperature to $0.2$--$0.3$ only for its intermediate query-generation steps and runs a default budget of \agentRounds{} rounds. Per-stage decoding, retrieval, and forecast-parsing settings are listed in Appendix~\ref{sec:appendix-config}.

A typical trace opens with a parallel batch of seed queries against the archive, expands the highest-ranked submissions with their full text and the author's recent history, and closes with the labeling and final-forecast passes. The prompts driving both pipelines are in Appendix~\ref{sec:appendix-pipelines}.

\providecommand{\stdev}[1]{\textsubscript{\tiny\,$\pm$#1}}
\definecolor{covbg}{gray}{0.94}
\providecommand{\blocksep}{\hspace{2.5pt}{\color{gray!60}\vrule width 0.6pt}\hspace{2.5pt}}
\begin{table*}[!htbp]
\footnotesize
\centering
\setlength{\tabcolsep}{2pt}
\renewcommand{\arraystretch}{1.12}
\resizebox{\textwidth}{!}{%
\begin{tabular}{@{}l r >{\columncolor{covbg}}r >{\columncolor{covbg}}r >{\columncolor{covbg}}r r !{\blocksep} r >{\columncolor{covbg}}r >{\columncolor{covbg}}r >{\columncolor{covbg}}r r@{}}
\toprule
                    & \multicolumn{5}{c!{\blocksep}}{\textbf{Zero-shot (Vanilla)}}
                    & \multicolumn{5}{c}{\textbf{Retrieval (Agent)}} \\
\cmidrule(lr){2-6}\cmidrule(lr){7-11}
                    &                              & \multicolumn{3}{c}{\cellcolor{covbg}\footnotesize\textit{Acc by coverage}} &                                  &                              & \multicolumn{3}{c}{\cellcolor{covbg}\footnotesize\textit{Acc by coverage}} &                                  \\
\cmidrule(lr){3-5}\cmidrule(lr){8-10}
\textbf{Model}      & {Acc\,$\uparrow$}            & {S}                  & {W}                  & {N}                  & {Brier\,$\downarrow$}            & {Acc\,$\uparrow$}            & {S}                  & {W}                  & {N}                  & {Brier\,$\downarrow$}            \\
\midrule
Qwen3-4B            & 60.1\stdev{1.0}              & 45.9\stdev{1.9}      & 64.9\stdev{3.3}      & 63.2\stdev{0.9}      & .245\stdev{.004}                 & \textbf{65.9}\stdev{2.1}     & 73.5\stdev{6.5}      & 70.3\stdev{4.6}      & 58.3\stdev{2.0}      & \textbf{.218}\stdev{.006}        \\
Qwen3-8B            & 55.8\stdev{0.4}              & 61.7\stdev{2.8}      & 54.7\stdev{3.2}      & 50.3\stdev{2.5}      & .261\stdev{.004}                 & \textbf{64.7}\stdev{2.2}     & 80.9\stdev{4.4}      & 63.7\stdev{6.2}      & 48.9\stdev{5.2}      & \textbf{.213}\stdev{.008}        \\
Qwen3-32B           & 64.6\stdev{1.0}              & 65.6\stdev{3.2}      & 67.9\stdev{6.2}      & 51.3\stdev{9.1}      & .234\stdev{.007}                 & \textbf{72.8}\stdev{3.0}     & 79.3\stdev{5.4}      & 70.5\stdev{3.6}      & 57.8\stdev{6.0}      & \textbf{.179}\stdev{.008}        \\
R1-Distill-Qwen-7B  & \textbf{58.7}\stdev{1.2}     & 54.2\stdev{3.6}      & 57.9\stdev{2.2}      & 52.2\stdev{3.8}      & \textbf{.256}\stdev{.012}        & 51.2\stdev{5.1}              & 50.9\stdev{6.8}      & 52.2\stdev{6.6}      & 48.5\stdev{7.6}      & .318\stdev{.015}                 \\
\midrule
Llama-3.2-3B        & 56.1\stdev{2.1}              & 53.7\stdev{9.4}      & 43.7\stdev{14.2}     & 53.0\stdev{1.4}      & .375\stdev{.013}                 & \textbf{56.5}\stdev{1.6}     & 51.8\stdev{13.3}     & 68.9\stdev{26.8}     & 56.7\stdev{1.3}      & \textbf{.292}\stdev{.006}        \\
Llama-3.1-8B        & \textbf{66.1}\stdev{1.0}     & 64.7\stdev{6.2}      & 67.7\stdev{2.1}      & 57.5\stdev{9.1}      & .250\stdev{.006}                 & 60.0\stdev{1.9}              & 78.3\stdev{10.1}     & 60.8\stdev{1.4}      & 40.5\stdev{7.9}      & \textbf{.230}\stdev{.010}        \\
\midrule
Gemma-3-4B          & 54.5\stdev{0.2}              & 53.8\stdev{6.0}      & 54.1\stdev{3.7}      & 47.7\stdev{41.2}     & .297\stdev{.002}                 & \textbf{56.2}\stdev{2.5}     & 56.6\stdev{3.2}      & 56.1\stdev{2.4}      & 60.3\stdev{27.2}     & \textbf{.262}\stdev{.008}        \\
Gemma-3-27B         & 70.5\stdev{0.2}              & 70.6\stdev{2.2}      & 72.1\stdev{1.5}      & 63.6\stdev{3.7}      & .214\stdev{.001}                 & \textbf{71.1}\stdev{3.2}     & 87.2\stdev{4.7}      & 68.7\stdev{4.5}      & 48.3\stdev{5.2}      & \textbf{.184}\stdev{.008}        \\
\midrule
Ministral-8B        & 52.9\stdev{1.2}              & 42.1\stdev{12.1}     & 60.0\stdev{3.0}      & 48.2\stdev{4.7}      & .281\stdev{.004}                 & \textbf{56.8}\stdev{2.9}     & 66.3\stdev{4.9}      & 59.0\stdev{5.2}      & 48.3\stdev{3.8}      & \textbf{.242}\stdev{.010}        \\
\bottomrule
\end{tabular}%
}
\caption{\textbf{Main results on the \evalN{}-market eval set at the
cutoff $t_0$ of \S\ref{sec:eval-metrics}.} Overall accuracy ($\uparrow$, in \%),
accuracy stratified by the coverage probe's three labels
(\emph{S}/\emph{W}/\emph{N}, shaded), and Brier ($\downarrow$, in
points). The horizontal rules separate model families (Qwen, Llama~3,
Gemma~3, Mistral). \emph{Zero-shot} is the bare model and
\emph{Retrieval} is the same model with archive access at $t_0$.
Subscripts are run-to-run standard deviation across five consistency
runs. Because the S/W/N bin assignment is per-run from the agent's
own retrieval tally, the per-bin std also reflects bin-membership
churn. Best Acc and best Brier per model are bolded.}
\label{tab:headline-results}
\vspace{-1\baselineskip}
\end{table*}

\section{Experiments}
\label{sec:experiments}

We benchmark \nModels{} open-weight LLMs under both pipelines on the eval set of \S\ref{sec:eval-construction} at the cutoff $t_0$ of \S\ref{sec:eval-metrics}, spanning the Qwen3 family at 4B, 8B, and 32B parameters~\citep{yang2025qwen3}, the Llama 3 family with Llama-3.1-8B-Instruct and Llama-3.2-3B-Instruct~\citep{dubey2024llama3}, Gemma~3 at 4B and 27B~\citep{kamath2025gemma3}, Ministral-8B-Instruct~\citep{mistral2024ministral}, and the DeepSeek-R1 distillation onto Qwen-7B~\citep{guo2025deepseekr1}.

All models share the prompt template, the decoding configuration of \S\ref{sec:pipelines-agent}, and the retrieval defaults, and all of them predate the start of the eval window. Any cross-family or cross-model variation therefore isolates either pretraining differences or the contribution of retrieval. Each (model, pipeline) cell is averaged over five consistency runs.

Table~\ref{tab:headline-results} reports the main results. The Retrieval agent lowers Brier on eight of the nine models, with the largest relative reduction on Qwen3-32B ($.234 \to .179$, a $23\%$ drop) and the largest accuracy lift on Qwen3-8B ($55.8\% \to 64.7\%$). Conditioning on the archive helps most on markets the coverage probe rates \emph{strong}, where retrieval lifts accuracy on most backbones (median $+14$ points), while \emph{None}-bin accuracy moves little. R1-Distill-Qwen-7B is the only model whose overall Brier rises under retrieval, and \S\ref{sec:analysis-per-model} unpacks this case. Per-run and per-bin numbers behind Table~\ref{tab:headline-results} are in Appendix~\ref{sec:appendix-results}.

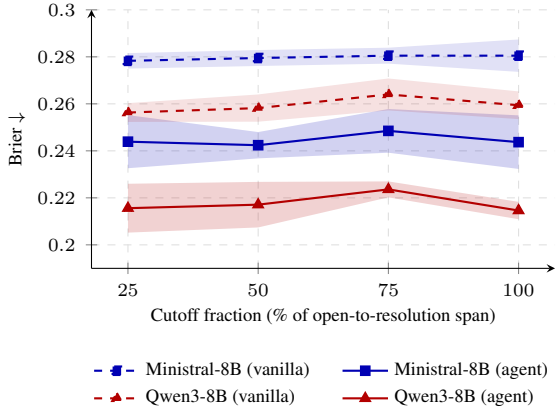
\begin{figure}[!b]
  \centering
  \begin{tikzpicture}
    \begin{axis}[
      width=\columnwidth,
      height=5.0cm,
      xlabel={Cutoff fraction (\% of open-to-resolution span)},
      ylabel={Brier $\downarrow$},
      xlabel style={font=\scriptsize, yshift=1mm},
      ylabel style={font=\scriptsize, yshift=-1mm},
      xtick={25,50,75,100},
      xmin=18, xmax=107,
      ymin=0.19, ymax=0.30,
      ytick={0.20, 0.22, 0.24, 0.26, 0.28, 0.30},
      tick label style={font=\scriptsize},
      tick align=outside,
      axis lines=left,
      grid=major,
      grid style={dashed, gray!25},
      legend style={
        font=\scriptsize,
        at={(0.5,-0.32)},
        anchor=north,
        legend columns=2,
        /tikz/every even column/.append style={column sep=0.35cm},
        draw=none,
        fill=none,
      },
      legend cell align={left},
    ]
      \addplot[name path=mvU, draw=none, forget plot]
        coordinates {(25,0.2816) (50,0.2829) (75,0.2839) (100,0.2874)};
      \addplot[name path=mvL, draw=none, forget plot]
        coordinates {(25,0.2750) (50,0.2761) (75,0.2771) (100,0.2736)};
      \addplot[blue!70!black, fill opacity=0.12, draw=none, forget plot]
        fill between[of=mvU and mvL];
      \addplot[mark=square*, mark size=1.6pt, color=blue!70!black, dashed, thick]
        coordinates {(25,0.2783) (50,0.2795) (75,0.2805) (100,0.2805)};
      \addlegendentry{Ministral-8B (vanilla)}

      \addplot[name path=maU, draw=none, forget plot]
        coordinates {(25,0.2552) (50,0.2480) (75,0.2578) (100,0.2551)};
      \addplot[name path=maL, draw=none, forget plot]
        coordinates {(25,0.2326) (50,0.2368) (75,0.2392) (100,0.2323)};
      \addplot[blue!70!black, fill opacity=0.18, draw=none, forget plot]
        fill between[of=maU and maL];
      \addplot[mark=square*, mark size=1.6pt, color=blue!70!black, thick]
        coordinates {(25,0.2439) (50,0.2424) (75,0.2485) (100,0.2437)};
      \addlegendentry{Ministral-8B (agent)}

      \addplot[name path=qvU, draw=none, forget plot]
        coordinates {(25,0.2603) (50,0.2640) (75,0.2708) (100,0.2653)};
      \addplot[name path=qvL, draw=none, forget plot]
        coordinates {(25,0.2523) (50,0.2524) (75,0.2572) (100,0.2535)};
      \addplot[red!70!black, fill opacity=0.12, draw=none, forget plot]
        fill between[of=qvU and qvL];
      \addplot[mark=triangle*, mark size=2.0pt, color=red!70!black, dashed, thick]
        coordinates {(25,0.2563) (50,0.2582) (75,0.2640) (100,0.2594)};
      \addlegendentry{Qwen3-8B (vanilla)}

      \addplot[name path=qaU, draw=none, forget plot]
        coordinates {(25,0.2260) (50,0.2268) (75,0.2270) (100,0.2183)};
      \addplot[name path=qaL, draw=none, forget plot]
        coordinates {(25,0.2052) (50,0.2074) (75,0.2202) (100,0.2109)};
      \addplot[red!70!black, fill opacity=0.18, draw=none, forget plot]
        fill between[of=qaU and qaL];
      \addplot[mark=triangle*, mark size=2.0pt, color=red!70!black, thick]
        coordinates {(25,0.2156) (50,0.2171) (75,0.2236) (100,0.2146)};
      \addlegendentry{Qwen3-8B (agent)}
    \end{axis}
  \end{tikzpicture}
  \caption{\textbf{Brier vs. per-market cutoff fraction for the
    duration sweep.} Each fraction places $t_0$ at $25\%$, $50\%$,
    $75\%$, or $100\%$ of an individual market's open-to-resolution
    span, so the per-market wall-clock cutoff scales with how long
    the market was actually open. The Retrieval agent (solid) lowers
    Brier over the zero-shot baseline (dashed) for both probed
    models, and the gap between them stays roughly constant across
    fractions. This flat profile is consistent with retrieval gain
    saturating well below the full available lookback. Points are
    means and shaded bands are $\pm 1$ standard deviation across five
    consistency runs per cell.}
  \label{fig:lookback-brier}
\end{figure}

\paragraph{Lookback fraction.}
The cutoff above is pinned to a wall-clock date, so it lands at a different point in each market's life depending on how long that market stayed open. As a finer probe of how much pre-resolution discussion the agent actually needs, we re-run a two-model subset at four per-market cutoff fractions, placing $t_0$ at $25\%$, $50\%$, $75\%$, and $100\%$ of each market's open-to-resolution span.

The zero-shot baseline barely moves across the four fractions for either probed model (Ministral-8B and Qwen3-8B), so any cross-fraction differences must come from the retrieval side. The Retrieval agent's accuracy and Brier also shift only slightly, with Qwen3-8B rising from $61.0\%$ accuracy at the $25\%$ fraction to $63.4\%$ at the $100\%$ fraction while its Brier hovers near $0.21$, and Ministral-8B's numbers essentially unchanged across the sweep. Figure~\ref{fig:lookback-brier} plots Brier against the cutoff fraction for both models and pipelines (per-cell numbers in Appendix~\ref{sec:appendix-lookback}), and the gap between the agent and its zero-shot baseline is roughly constant across fractions for each model. The agent's gain therefore saturates well below the full available lookback, which suggests the wall-clock offset choice is not picking up an artifact of how late in each market's life the cutoff lands.

\section{Analysis}
\label{sec:analysis}

Underneath the headline averages, the picture changes sharply from topic to topic. For every market we pair the Agent and Vanilla runs by id, pool the eight models that have per-run logs, and count, per topic, how often the agent fixes a market the zero-shot baseline got wrong (a \emph{recovery}) against how often it breaks one the baseline got right (a \emph{regression}). The net of the two swings hard in both directions from topic to topic, as Figure~\ref{fig:failure-topics} shows.

\subsection{Where retrieval helps}
\label{sec:analysis-helps}

The positive net is concentrated in Sports ($+16$), Awards ($+12$), and Trading/commodities ($+7$). What these topics have in common is that Reddit discusses them in concrete detail in the weeks before they resolve, so the agent's opening queries land on posts with real signal. They also resolve on facts anyone can check after the fact, like which team won, who took the award, or whether a price crossed a line, so an agent that has found the right thread can read the answer back. Both of these push the coverage probe toward \emph{strong}, and the four topics on the right of Figure~\ref{fig:failure-topics} account for most of the strong-bin accuracy lift in Table~\ref{tab:headline-results}. They are also where the agent quits early most often. Its confidence-driven stop fires on $52$ recovered markets against $34$ regressed ones, so once the evidence is in front of it the agent commits before spending its whole round budget.

\subsection{Where retrieval hurts}
\label{sec:analysis-hurts}

The damage is concentrated in Entertainment at $-31$, with Other ($-10$), Elections/politics ($-7$), Social/celebrity ($-7$), Tech/AI ($-5$), and Crypto ($-3$) trailing behind. Entertainment shows the failure most cleanly. Nearly all of its markets are Billboard ``will X be the \#1 song this week'' questions, and nearly all of them resolve \emph{No}, since most songs never reach the top of the Hot~100. A model with no retrieval just sits near that base rate and is usually right. The agent goes looking instead, finds the pre-release hype that music subreddits produce for any plausible contender, and counts that enthusiasm as evidence for Yes. Its tally and reflection steps reward lopsided evidence, but hype is lopsided too, and the scaffold cannot tell a fan's excitement from a fact about the chart. The agent argues itself out of a correct default. The milder regressions on Social/celebrity, Tech/AI, and Crypto are the same failure in lower volume, all places where the loudest posts are opinion that does not track the outcome. The duration split fits this reading. Long-lived markets tip toward regression, with $144$ regressions against $106$ recoveries, while medium-lived ones come out slightly positive, with $329$ regressions against $381$ recoveries, as though the longer a market stays open the more chatter accumulates for the agent to overfit.

\definecolor{barneg}{rgb}{0.72,0.30,0.28}
\definecolor{barpos}{rgb}{0.17,0.35,0.63}

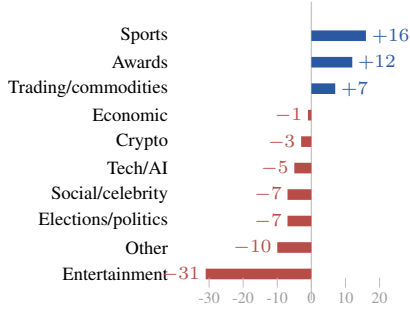
\begin{figure}[!t]
  \centering
  \begin{tikzpicture}[
    font=\scriptsize,
    bar/.style={line width=4pt, line cap=butt},
    posbar/.style={bar, draw=barpos},
    negbar/.style={bar, draw=barneg},
    valpos/.style={font=\scriptsize\bfseries, color=barpos},
    valneg/.style={font=\scriptsize\bfseries, color=barneg},
    topiclab/.style={anchor=east, inner sep=1pt, font=\scriptsize},
  ]
  \def\rowstep{0.35}
  \def\barscale{0.045}
  \def\labelw{2.45}
  \def\zerox{4.30}

  \draw[gray!45, line width=0.5pt]
    (\zerox, -0.35) -- (\zerox, 3.6);

  \foreach \i/\name in {%
    0/Entertainment,%
    1/Other,%
    2/{Elections/politics},%
    3/{Social/celebrity},%
    4/{Tech/AI},%
    5/Crypto,%
    6/Economic,%
    7/{Trading/commodities},%
    8/Awards,%
    9/Sports%
  }{
    \pgfmathsetmacro{\yc}{\i * \rowstep}
    \node[topiclab] at (\labelw, \yc) {\name};
  }

  \foreach \i/\v in {0/31, 1/10, 2/7, 3/7, 4/5, 5/3, 6/1} {
    \pgfmathsetmacro{\yc}{\i * \rowstep}
    \pgfmathsetmacro{\xend}{\zerox - \v * \barscale}
    \draw[negbar] (\zerox, \yc) -- (\xend, \yc);
    \node[valneg, anchor=east, inner sep=2pt] at (\xend, \yc) {$-\v$};
  }

  \foreach \i/\v in {7/7, 8/12, 9/16} {
    \pgfmathsetmacro{\yc}{\i * \rowstep}
    \pgfmathsetmacro{\xend}{\zerox + \v * \barscale}
    \draw[posbar] (\zerox, \yc) -- (\xend, \yc);
    \node[valpos, anchor=west, inner sep=2pt] at (\xend, \yc) {$+\v$};
  }

  \foreach \x in {-30,-20,-10,0,10,20} {
    \pgfmathsetmacro{\tx}{\zerox + \x * \barscale}
    \draw[gray!50, line width=0.3pt] (\tx, -0.18) -- (\tx, -0.05);
    \node[font=\tiny\color{gray!75}, anchor=north, inner sep=1.5pt]
      at (\tx, -0.18) {\x};
  }

  \end{tikzpicture}
  \caption{\textbf{Per-topic net effect of the Retrieval agent.}
    For each topic with at least four markets in the eval set, the
    bar shows the number of markets the agent \emph{recovers} (agent
    right, zero-shot baseline wrong) minus the number it
    \emph{regresses} (the reverse), summed across the eight models
    with per-run logs available. Blue bars right of zero are topics
    where retrieval helps on net, and red bars left of zero are
    topics where it hurts.}
  \label{fig:failure-topics}
  \vspace{-1\baselineskip}
\end{figure}

\subsection{Per-model effects, and accuracy versus calibration}
\label{sec:analysis-per-model}

Every model shows the same topic pattern, but the size of the effect varies widely. Qwen3-8B has the cleanest split at $+43$ markets, Qwen3-4B is solidly positive at $+19$, and Gemma-3-4B comes out even. The rest land negative, slightly for Llama-3.2-3B and Ministral-8B-Instruct ($-3$ and $-6$) and more clearly for Llama-3.1-8B-Instruct, Gemma-3-27B, and R1-Distill-Qwen-7B ($-27$, $-10$, and $-19$). The Qwen3 models pull far more from the same evidence stream than the others, which fits their strong-bin lift in Table~\ref{tab:headline-results}.

Accuracy and Brier do not always agree, since one counts decision flips and the other tracks where the probability mass ends up. Llama-3.1-8B-Instruct is the clearest example. It regresses on $91$ markets against only $64$ recoveries ($-27$ net), and its Brier still improves under retrieval ($.250 \to .230$). The agent flips more calls the wrong way than the baseline does, but it hedges those wrong calls, and Brier would rather see a wrong answer at $0.55$ than at $0.95$. Even where retrieval costs accuracy, then, it drags the probabilities back toward the middle and the calibration improves anyway. R1-Distill-Qwen-7B is the one model where both numbers move the wrong way at once. We read that as a context-budget problem, since its long reasoning chains crowd out the evidence while the same retrieval surface wins cleanly on Qwen3-8B at the same scale.

\section{Conclusion}
\label{sec:conclusion}

\sysname{} grades LLM forecasters at a past time $t_0$, pinning retrieval to an immutable pre-$t_0$ Reddit snapshot and scoring against both the resolved Polymarket outcome and the matched-time price. Retrieval lowers Brier on eight of nine models (up to 23\%), but the average hides a topic split that the cutoff reveals. Retrieval recovers Sports, Awards, and Trading, where Reddit carries fact-anchored pre-event discussion, and regresses on Entertainment, where speculative hype reads as evidence. A retrieval-grounded forecaster is only as good as the signal the corpus already carried before resolution.

The design also gives the evaluation levers that live benchmarks lack. The cutoff can be swept along each market's life to trace how calibration develops as evidence accumulates, and new predictors can be dropped behind the same interface and replayed on identical evidence. New markets keep resolving against the same frozen archive, so the test stays ahead of each new model's training cutoff.

\section*{Limitations}

The archive is built from a fixed set of \corpusSubreddits{} public subreddits, so the corpus skews English-language and community-driven and under-represents topics Reddit discusses thinly (e.g., regional politics, niche corporate events). Polymarket inherits a similar skew, and we restrict to binary Yes/No markets, leaving multi-outcome and continuous markets out of scope.

Because the archive ends on 2026-01-31, most markets in our eval set have $t_0$ pinned at the archive boundary with only a short pre-resolution lookback (median close--$t_0$ gap of 23 days), which limits what we can say about long-horizon forecasting. We also benchmark only \nModels{} open-weight models behind two reference pipelines, and closed-weight models and alternative agent scaffolds are left to future work.

\sysname{} is built from public Reddit content and historical Polymarket data. The privacy, deployment, and content risks this raises, and how we address them, are discussed in the Ethics Statement below.

\section*{Ethics Statement}

Our corpus is derived from the public Pushshift Reddit dataset and our questions from historical Polymarket data. Both are public resources used strictly for non-commercial research, and the posts remain subject to Reddit's User Agreement. The benchmarked checkpoints are used within their licenses (Apache-2.0 for the Qwen3 models and the Qwen3-Embedding encoder; the Llama 3.1/3.2 Community Licenses; the Gemma Terms of Use; MIT for DeepSeek-R1-Distill-Qwen-7B; and the research-only Mistral Research License for Ministral-8B-Instruct-2410), and all use here is research-only and consistent with these terms.

The corpus contains only public posts. We retain pseudonymous author handles for thread- and author-level lookups but make no attempt at re-identification, drop documents whose author or body is deleted or removed and accounts on a bot list, and release document and market identifiers with regeneration code, keeping raw post text out of the release. Artifacts derived from the corpus should not be used to profile or de-anonymize individuals. We do not screen for offensive content. Because the filters condition only on engagement and length, some posts may carry biased or offensive language that an evaluated model could echo.

Finally, the predictors we study are forecasting agents. Deployed without calibration audits they could lend false authority to automated bets or decisions, and our analysis shows that retrieval can make a model confidently wrong on speculative topics. We therefore frame \sysname{} as an evaluation testbed and caution against deploying it as a trading tool. All released artifacts are intended for research use only.

\bibliography{custom}

\appendix
\section*{Appendix Contents}
\label{sec:appendix-toc}
\begingroup
\setlength{\parskip}{2pt}
\noindent
\textbf{\ref{sec:appendix-backend}~\nameref{sec:appendix-backend}}\\
\hspace*{1em}\ref{sec:appendix-tools}~\nameref{sec:appendix-tools}\\
\hspace*{1em}\ref{sec:appendix-latency}~\nameref{sec:appendix-latency}\\
\textbf{\ref{sec:appendix-pipelines}~\nameref{sec:appendix-pipelines}}\\
\hspace*{1em}\ref{sec:appendix-prompts-vanilla}~\nameref{sec:appendix-prompts-vanilla}\\
\hspace*{1em}\ref{sec:appendix-prompts-agent}~\nameref{sec:appendix-prompts-agent}\\
\hspace*{1em}\ref{sec:appendix-config}~\nameref{sec:appendix-config}\\
\textbf{\ref{sec:appendix-results}~\nameref{sec:appendix-results}}\\
\hspace*{1em}\ref{sec:appendix-per-run}~\nameref{sec:appendix-per-run}\\
\hspace*{1em}\ref{sec:appendix-bin-sizes}~\nameref{sec:appendix-bin-sizes}\\
\hspace*{1em}\ref{sec:appendix-lookback}~\nameref{sec:appendix-lookback}\\
\textbf{\ref{sec:appendix-ai}~\nameref{sec:appendix-ai}}
\endgroup
\vspace{0.5\baselineskip}

\section{Backend Details}
\label{sec:appendix-backend}

\subsection{Tool Surface}
\label{sec:appendix-tools}
Table~\ref{tab:tools} lists the five tools exposed to the Retrieval
agent. All five accept the same filter block (cutoff $t_0$, document
type, optional subreddit, author, month, and start-time bounds) and
inherit the row-level cutoff guarantee of \S\ref{sec:backend-cutoff}.

\begin{table}[t]
\small
\centering
\begin{tabular}{@{}lp{0.55\columnwidth}@{}}
\toprule
\textbf{Tool} & \textbf{Purpose} \\
\midrule
\texttt{search\_database}     & Dense (vector) retrieval over submissions. \\
\texttt{getpostcoreinfo}      & Fetch a post by id. \\
\texttt{getcommentcoreinfo}   & Fetch a comment by id. \\
\texttt{getpostcommentslist}  & List comments under a post, with optional ancestor / descendant thread traversal. \\
\texttt{getauthorhistorylist} & List recent posts and comments by an author. \\
\bottomrule
\end{tabular}
\caption{The five tools exposed to the prediction agent. All accept the same filter block and inherit the cutoff guarantee (\S\ref{sec:backend-cutoff}).}
\label{tab:tools}
\end{table}

\subsection{Retrieval Latency}
\label{sec:appendix-latency}

The partition selector of \S\ref{sec:backend-cutoff} restricts each
dense query to the monthly partitions admitted by the cutoff $t_0$,
so retrieval cost tracks the number of partitions admitted by the
cutoff, independent of total corpus size. Table~\ref{tab:cutoff-latency}
measures this against the live index. Per-query latency rises
monotonically with the admitted partition count, from a median of
$0.14$\,s at three partitions to $0.84$\,s at all \corpusMonths{}.
The cutoff used in our experiments, the archive's final day
(2026-01-31), sits at the top of this range, since it admits
nearly every partition. Latencies are end-to-end per query,
comprising query embedding, per-partition HNSW search, and merge.
They are measured over the running retrieval server on a single GPU
and averaged over 32 queries per cutoff, so absolute values will
vary with hardware and load.

\begin{table}[t]
\centering
\small
\begin{tabular}{@{}rrr@{}}
\toprule
Partitions $\le t_0$ & Median (ms) & p95 (ms) \\
\midrule
\phantom{0}3  & 144 & 155 \\
\phantom{0}9  & 392 & 410 \\
15 & 516 & 555 \\
21 & 710 & 786 \\
25 & 836 & 944 \\
\bottomrule
\end{tabular}
\caption{Per-query dense-retrieval latency against the live index as a function of the monthly partitions admitted by the cutoff $t_0$. The end-to-end measurement covers query embedding, per-partition HNSW search, and merge on a single GPU, averaged over 32 queries per row. Cost scales with the partitions admitted by the cutoff, so earlier cutoffs are cheaper. Most markets in our eval set pin $t_0$ at the archive boundary (2026-01-31), so retrieval admits essentially all \corpusMonths{} partitions (bottom row).}
\label{tab:cutoff-latency}
\end{table}

\section{Pipeline Details}
\label{sec:appendix-pipelines}

This appendix gives the prompts and per-call configuration that
instantiate the two pipelines of \S\ref{sec:pipelines}. The tool
surface they call through is documented separately in
Appendix~\ref{sec:appendix-tools}.

\subsection{Zero-shot (Vanilla) Prompt}
\label{sec:appendix-prompts-vanilla}
The vanilla pipeline has no retrieval. The model receives the
system line below, followed by the per-market user template.

\paragraph{System.}
\begingroup
\footnotesize
\begin{verbatim}
You are a forecaster. Predict the probability
of the event.
\end{verbatim}
\endgroup

\paragraph{User template.}
\begingroup
\footnotesize
\begin{verbatim}
Question: {question}

Resolution criteria:
{description}

Forecast cutoff date: {cutoff_date}

Output a single JSON object only, matching:
{
  "choice": "Yes" | "No",
  "prob_yes": 0.0-1.0,
  "prob_no": 0.0-1.0,
  "confidence": 0.0-1.0,
  "reasoning": "brief"
}
\end{verbatim}
\endgroup

\subsection{Retrieval-Agent Prompts}
\label{sec:appendix-prompts-agent}
The agent pipeline runs in two stages. A research stage drives the
ReAct tool loop and writes an evidence report. A forecasting stage
turns that report into a calibrated forecast under the same schema
as the vanilla pipeline. Both stages share a cutoff date set to
the simulated query time $t_0$.

\paragraph{Research system.}
\begingroup
\footnotesize
\begin{verbatim}
You are preparing evidence for a forecasting
task.

Use ONLY the local Reddit tools. Do not use
web search, browser tools, or outside
knowledge as evidence. Every search must
respect the provided cutoff date. Gather
broad evidence first, then follow up on
concrete names, dates, organizations, IDs,
and claims found in the first pass.
\end{verbatim}
\endgroup

\paragraph{Research user template (abridged).}
\begingroup
\footnotesize
\begin{verbatim}
Cutoff date for all tool calls: {cutoff_date}

Task:
{prompt}

Options:
{options}

Research requirements:
- Generate multiple diverse research
  questions instead of searching the
  question verbatim.
- Use many local Reddit tool calls: broad
  vector searches first, then targeted
  follow-ups.
- Retrieve full post/comment/thread/author
  context when search hits look relevant.
- Produce an evidence report with sections:
  Key facts, Evidence for each option,
  Evidence against each option, Uncertainty,
  and Suggested probability direction.
- Do not use information after the cutoff
  date.

Local Reddit tool rules:
- Available subreddits are exactly:
  {AVAILABLE_SUBREDDITS}
- search_database must use
  doc_type="submission" only.
- Valid month format is "YYYY-MM" only.
\end{verbatim}
\endgroup

\paragraph{Forecast system.}
\begingroup
\footnotesize
\begin{verbatim}
You are a calibrated forecasting judge.

You receive a Polymarket-style question plus
a local Reddit evidence report. Your job is
to produce one forced-choice forecast with
calibrated probabilities.

Forecasting approach:
1. Form your own base-rate prior using world
   knowledge, the resolution criteria, and
   the timing/cutoff.
2. Compare the strength and asymmetry of
   evidence on each side. Move away from the
   prior in proportion to that asymmetry;
   confident moves require asymmetric
   evidence, not absence of evidence.
3. Treat indirect/contextual signals as
   actionable when they consistently lean
   one way.

Return ONLY a single JSON object matching
the requested schema.
\end{verbatim}
\endgroup

\paragraph{Forecast user template (abridged).}
Step 1 sets a prior anchor based on a five-category classification
of the question. Step 2 walks the model through tally-based updates
from the evidence report. Step 3 emits the same JSON schema as the
vanilla pipeline.
\begingroup
\footnotesize
\begin{verbatim}
Question: {question}
Resolution criteria: {description}
Allowed options: {options}
Forecast cutoff date: {cutoff_date}

Step 1 - PRIOR (ignore Reddit evidence for
this step). Pick a specific anchor from one
of the five categories below:
  1. Narrow numeric range       -> .05-.20
  2. Underdog / against-trend   -> .10-.30
  3. Structurally uncertain     -> .40-.60
  4. Favorite / with-trend      -> .55-.85
  5. Broadly likely outcomes    -> .70-.95
State this prior P(Yes) as the first
sentence of "reasoning".

Step 2 - ADJUST using the Reddit evidence:
{report}

- Clear asymmetry: move the prior toward
  the stronger side.
- evidence_quality is "none" or "weak" with
  no asymmetry: keep the Step-1 prior,
  unless it was extreme and anchor-driven.
- Indirect/contextual signals ARE actionable
  when they consistently lean one way.

Step 3 - Output the final forecast as a
single JSON object matching the requested
schema.
\end{verbatim}
\endgroup

\subsection{Decoding and Retrieval Configuration}
\label{sec:appendix-config}

Table~\ref{tab:pipeline-config} lists the per-call decoding settings. Top-$p$ is left at the vLLM default ($1.0$) throughout. Thinking is left at each model's default (on for the Qwen3 family), and any \texttt{<think>} span is stripped from a reply before it is parsed.

\begin{table}[t]
\centering
\small
\begin{tabular}{@{}lcr@{}}
\toprule
LLM call & Temp. & Max tokens \\
\midrule
Query generation        & 0.2 & 4096 \\
Follow-up / reflection   & 0.3 & 4096 \\
Evidence summarization   & 0.1 & 3500 \\
Evidence tally           & 0.1 & 2200 \\
Final forecast           & 0.1 & 3200 \\
Forecast retry          & 0.1 & 1400/2400/900 \\
Zero-shot baseline       & 0.1 & 4096 \\
\bottomrule
\end{tabular}
\caption{Per-call decoding settings for the two pipelines. Top-$p$ is left at the vLLM default ($1.0$) throughout. The agent forecasts at temperature $0.1$ and raises the temperature only for the intermediate query-generation and reflection steps that benefit from diversity. The three forecast-retry budgets ($1400/2400/900$) are the repair, compact, and JSON-only fallback passes.}
\label{tab:pipeline-config}
\end{table}

\paragraph{Retrieval.}
The agent runs a default budget of \agentRounds{} rounds. Round one issues six seed queries, and each subsequent round adds four to six follow-up queries. All searches are dense (vector) over submissions (\S\ref{sec:backend-search}). The backend returns the top $20$ hits per query, and the agent expands the $20$ highest-ranked submissions with their full post text and the author's recent history (up to five posts and five comments each). The loop stops early when the running confidence reaches $0.7$ or the round budget is spent.

\paragraph{Forecast parsing.}
The final forecast is read from a JSON object through a four-step fallback. We first parse the main forecast, and if that fails we run a JSON-repair pass, then a compact re-prompt, and finally a strict JSON-only re-prompt. Across all \nModels{} models this leaves at most three of \evalN{} markets without a parsed prediction in a run, so truncation of the forecast pass is not a material source of missing data.

\section{Full Results}
\label{sec:appendix-results}

\subsection{Per-Run Numbers}
\label{sec:appendix-per-run}
Table~\ref{tab:per-run} reports accuracy and Brier for all five
consistency runs behind each cell of Table~\ref{tab:headline-results}.

\begin{table*}[t]
\footnotesize
\centering
\setlength{\tabcolsep}{4pt}
\renewcommand{\arraystretch}{1.05}
\begin{tabular}{@{}l l rr rr rr rr rr@{}}
\toprule
                    &           & \multicolumn{2}{c}{\textbf{Run 1}} & \multicolumn{2}{c}{\textbf{Run 2}} & \multicolumn{2}{c}{\textbf{Run 3}} & \multicolumn{2}{c}{\textbf{Run 4}} & \multicolumn{2}{c}{\textbf{Run 5}} \\
\cmidrule(lr){3-4}\cmidrule(lr){5-6}\cmidrule(lr){7-8}\cmidrule(lr){9-10}\cmidrule(lr){11-12}
\textbf{Model}      & Pipe      & Acc   & Brier  & Acc   & Brier  & Acc   & Brier  & Acc   & Brier  & Acc   & Brier  \\
\midrule
Qwen3-4B            & Vanilla   & 59.8  & .238   & 60.5  & .247   & 59.0  & .247   & 59.7  & .247   & 61.6  & .245   \\
                    & Agent     & 68.8  & .212   & 66.1  & .226   & 66.7  & .212   & 64.2  & .220   & 63.7  & .223   \\
\midrule
Llama-3.2-3B        & Vanilla   & 53.0  & .396   & 57.9  & .367   & 54.9  & .377   & 57.4  & .362   & 57.3  & .374   \\
                    & Agent     & 56.5  & .285   & 54.9  & .296   & 55.1  & .296   & 58.6  & .296   & 57.4  & .286   \\
\midrule
Qwen3-8B            & Vanilla   & 56.1  & .260   & 55.2  & .263   & 55.6  & .265   & 55.9  & .254   & 56.3  & .263   \\
                    & Agent     & 66.1  & .199   & 62.5  & .215   & 62.0  & .221   & 66.7  & .216   & 66.1  & .213   \\
\midrule
Ministral-8B        & Vanilla   & 53.7  & .279   & 54.6  & .275   & 51.9  & .281   & 51.9  & .283   & 52.3  & .287   \\
                    & Agent     & 60.7  & .225   & 56.9  & .244   & 58.3  & .241   & 53.7  & .252   & 54.2  & .245   \\
\midrule
Llama-3.1-8B        & Vanilla   & 66.5  & .253   & 66.2  & .248   & 66.1  & .245   & 64.4  & .259   & 67.1  & .245   \\
                    & Agent     & 56.9  & .246   & 59.3  & .224   & 61.1  & .233   & 61.6  & .224   & 61.1  & .222   \\
\midrule
R1-Distill-Qwen-7B  & Vanilla   & 58.7  & .267   & 57.8  & .236   & 60.5  & .258   & 59.2  & .257   & 57.5  & .261   \\
                    & Agent     & 54.2  & .321   & 58.3  & .299   & 50.0  & .309   & 45.4  & .336   & 48.2  & .327   \\
\midrule
Gemma-3-4B          & Vanilla   & 54.2  & .301   & 54.6  & .296   & 54.6  & .296   & 54.6  & .297   & 54.6  & .296   \\
                    & Agent     & 60.2  & .248   & 54.6  & .269   & 53.7  & .263   & 56.0  & .264   & 56.5  & .263   \\
\midrule
Gemma-3-27B         & Vanilla   & 70.4  & .215   & 70.4  & .215   & 70.8  & .213   & 70.4  & .214   & 70.4  & .213   \\
                    & Agent     & 69.9  & .180   & 75.5  & .176   & 66.7  & .196   & 72.2  & .180   & 71.3  & .186   \\

\midrule
Qwen3-32B           & Vanilla   & 63.0  & .239   & 64.8  & .222   & 64.8  & .236   & 65.3  & .236   & 65.3  & .236   \\
                    & Agent     & 72.7  & .176   & 75.5  & .170   & 69.0  & .190   & 70.8  & .183   & 75.9  & .174   \\
\bottomrule
\end{tabular}
\caption{\textbf{Per-run accuracy and Brier for all five consistency
runs.} Same models and pipelines as Table~\ref{tab:headline-results}.
This is the data the run-to-run subscripts there are computed over.}
\label{tab:per-run}
\end{table*}

\subsection{Coverage Bin Sizes}
\label{sec:appendix-bin-sizes}
Table~\ref{tab:bin-sizes} reports the mean coverage bin sizes per
model (S/W/N), averaged across the five Agent runs that produced
them.

\definecolor{covbg}{gray}{0.94}
\begin{table}[t]
\footnotesize
\centering
\setlength{\tabcolsep}{4pt}
\renewcommand{\arraystretch}{1.08}
\begin{tabular}{@{}l >{\columncolor{covbg}}r >{\columncolor{covbg}}r >{\columncolor{covbg}}r r@{}}
\toprule
\textbf{Model}      & {S}   & {W}   & {N}   & {Total} \\
\midrule
Qwen3-4B            & 55    & 63    & 98    & 216 \\
Llama-3.2-3B        & 25    & 6     & 185   & 216 \\
Qwen3-8B            & 65    & 86    & 65    & 216 \\
Ministral-8B        & 32    & 116   & 68    & 216 \\
Llama-3.1-8B        & 20    & 170   & 26    & 216 \\
R1-Distill-Qwen-7B  & 60    & 113   & 40    & 213 \\
Gemma-3-4B          & 75    & 137   & 4     & 216 \\
Gemma-3-27B         & 62    & 121   & 33    & 216 \\
Qwen3-32B           & 100    & 87   & 29    & 216 \\
\bottomrule
\end{tabular}
\caption{\textbf{Coverage bin sizes (mean over 5 Agent runs).} The
S/W/N label is assigned per run from that run's own retrieval tally,
so a single market can land in different bins across runs. Sizes also
vary widely across models. Llama-3.2-3B reaches \emph{strong} on only
25 markets, and Gemma-3-4B reaches \emph{none} on only 4, which is
why Gemma-3-4B's \emph{N} cell in Table~\ref{tab:headline-results}
has very high std. R1-Distill-Qwen-7B totals 213
because three markets had no parseable retrieval tally in any of the
five runs.}
\label{tab:bin-sizes}
\end{table}

\subsection{Lookback-Fraction Sweep}
\label{sec:appendix-lookback}
\begin{figure}[t]
  \centering
  \begin{tikzpicture}
    \begin{axis}[
      width=\columnwidth,
      height=5.0cm,
      xlabel={Cutoff fraction (\% of open-to-resolution span)},
      ylabel={Accuracy $\uparrow$ (\%)},
      xlabel style={font=\scriptsize, yshift=1mm},
      ylabel style={font=\scriptsize, yshift=-1mm},
      xtick={25,50,75,100},
      xmin=18, xmax=107,
      ymin=50, ymax=67,
      ytick={50, 55, 60, 65},
      tick label style={font=\scriptsize},
      tick align=outside,
      axis lines=left,
      grid=major,
      grid style={dashed, gray!25},
      legend style={
        font=\scriptsize,
        at={(0.5,-0.32)},
        anchor=north,
        legend columns=2,
        /tikz/every even column/.append style={column sep=0.35cm},
        draw=none,
        fill=none,
      },
      legend cell align={left},
    ]
      \addplot[name path=mvU, draw=none, forget plot]
        coordinates {(25,54.4) (50,54.1) (75,54.3) (100,55.2)};
      \addplot[name path=mvL, draw=none, forget plot]
        coordinates {(25,52.8) (50,52.5) (75,52.1) (100,51.6)};
      \addplot[blue!70!black, fill opacity=0.12, draw=none, forget plot]
        fill between[of=mvU and mvL];
      \addplot[mark=square*, mark size=1.6pt, color=blue!70!black, dashed, thick]
        coordinates {(25,53.6) (50,53.3) (75,53.2) (100,53.4)};
      \addlegendentry{Ministral-8B (vanilla)}

      \addplot[name path=maU, draw=none, forget plot]
        coordinates {(25,60.7) (50,57.9) (75,57.7) (100,60.4)};
      \addplot[name path=maL, draw=none, forget plot]
        coordinates {(25,53.1) (50,56.1) (75,52.9) (100,52.6)};
      \addplot[blue!70!black, fill opacity=0.18, draw=none, forget plot]
        fill between[of=maU and maL];
      \addplot[mark=square*, mark size=1.6pt, color=blue!70!black, thick]
        coordinates {(25,56.9) (50,57.0) (75,55.3) (100,56.5)};
      \addlegendentry{Ministral-8B (agent)}

      \addplot[name path=qvU, draw=none, forget plot]
        coordinates {(25,57.3) (50,58.0) (75,56.5) (100,56.9)};
      \addplot[name path=qvL, draw=none, forget plot]
        coordinates {(25,55.5) (50,54.6) (75,53.1) (100,54.7)};
      \addplot[red!70!black, fill opacity=0.12, draw=none, forget plot]
        fill between[of=qvU and qvL];
      \addplot[mark=triangle*, mark size=2.0pt, color=red!70!black, dashed, thick]
        coordinates {(25,56.4) (50,56.3) (75,54.8) (100,55.8)};
      \addlegendentry{Qwen3-8B (vanilla)}

      \addplot[name path=qaU, draw=none, forget plot]
        coordinates {(25,63.3) (50,65.5) (75,64.4) (100,65.6)};
      \addplot[name path=qaL, draw=none, forget plot]
        coordinates {(25,58.7) (50,60.7) (75,60.8) (100,61.2)};
      \addplot[red!70!black, fill opacity=0.18, draw=none, forget plot]
        fill between[of=qaU and qaL];
      \addplot[mark=triangle*, mark size=2.0pt, color=red!70!black, thick]
        coordinates {(25,61.0) (50,63.1) (75,62.6) (100,63.4)};
      \addlegendentry{Qwen3-8B (agent)}
    \end{axis}
  \end{tikzpicture}
  \caption{\textbf{Accuracy vs. per-market cutoff fraction for the
    duration sweep.} Accuracy companion to the Brier plot in
    Figure~\ref{fig:lookback-brier}, over the same two-model subset,
    the same cutoff fractions, and the same band convention. The
    Retrieval agent (solid) lifts accuracy over the zero-shot
    baseline (dashed) for both models, and the gap stays roughly
    constant across fractions, matching the Brier picture.}
  \label{fig:lookback-acc}
\end{figure}
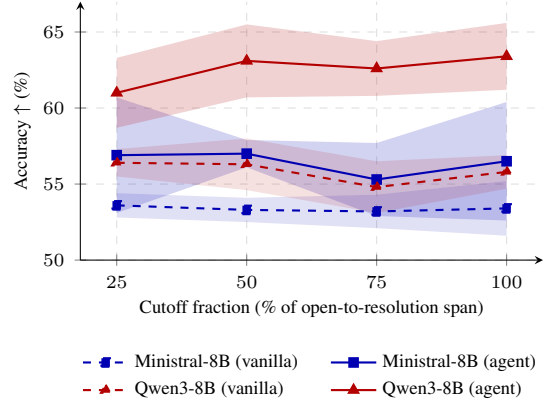

Table~\ref{tab:lookback-fraction} reports the per-cell accuracy and
Brier for the lookback-fraction sweep of \S\ref{sec:experiments},
i.e.\ the data plotted in Figure~\ref{fig:lookback-brier}.
Figure~\ref{fig:lookback-acc} plots the accuracy half of the same
sweep, which tracks the Brier picture.

\begin{table}[t]
\footnotesize
\centering
\setlength{\tabcolsep}{4pt}
\renewcommand{\arraystretch}{1.05}
\begin{tabular}{@{}llrrrr@{}}
\toprule
                  &              & \multicolumn{2}{c}{\textbf{Vanilla}} & \multicolumn{2}{c}{\textbf{Agent}} \\
\cmidrule(lr){3-4}\cmidrule(lr){5-6}
\textbf{Model}    & \textbf{Frac.} & Acc            & Brier            & Acc            & Brier            \\
\midrule
Ministral-8B      & \phantom{0}25\%  & 53.6 & .278 & 56.9 & .244 \\
                  & \phantom{0}50\%  & 53.3 & .280 & 57.0 & .242 \\
                  & \phantom{0}75\%  & 53.2 & .281 & 55.3 & .249 \\
                  & 100\%            & 53.4 & .281 & 56.5 & .244 \\
\midrule
Qwen3-8B          & \phantom{0}25\%  & 56.4 & .256 & 61.0 & .216 \\
                  & \phantom{0}50\%  & 56.3 & .258 & 63.1 & .217 \\
                  & \phantom{0}75\%  & 54.8 & .264 & 62.6 & .224 \\
                  & 100\%            & 55.8 & .259 & 63.4 & .215 \\
\bottomrule
\end{tabular}
\caption{\textbf{Lookback-fraction sweep on the two-model subset.}
Mean accuracy ($\uparrow$, in \%) and Brier ($\downarrow$, in points)
over five consistency runs per cell, for the per-market cutoff
fractions of \S\ref{sec:experiments} (placing $t_0$ at $25\%$,
$50\%$, $75\%$, or $100\%$ of each market's open-to-resolution span).
These are the numbers underlying Figure~\ref{fig:lookback-brier}.}
\label{tab:lookback-fraction}
\end{table}

\section{Use of AI Assistants}
\label{sec:appendix-ai}
We used a large language model solely as a writing aid, to polish the wording of the prose. It was not used for research ideation, experimental design, code generation, data analysis, or interpreting results, and all technical content and claims are the authors' own. The authors reviewed and verified all text and take full responsibility for the final manuscript.

\end{document}